\documentclass[conference]{IEEEtran}
\IEEEoverridecommandlockouts
\usepackage{multirow}
\usepackage{cite}
\usepackage{amsmath,amssymb,amsfonts}
\usepackage{algorithmic}
\usepackage{graphicx}
\usepackage{textcomp}
\usepackage{xcolor}
\usepackage{url}
\usepackage{comment}
\def\BibTeX{{\rm B\kern-.05em{\sc i\kern-.025em b}\kern-.08em
    T\kern-.1667em\lower.7ex\hbox{E}\kern-.125emX}}
\begin{document}

\title{inLoRA-MINT: Internal Representations for \\Auditing Training Data in Domain-adapted LLMs 
\thanks{Support by M2RAI (PID2024-160053OB-I00, MICIU/FEDER) and Cátedra ENIA UAM-VERIDAS en IA Responsable (NextGenerationEU PRTR TSI-100927-2023-2). Mancera is supported by FPI-PRE2022-104499 MICINN/FEDER and DeAlcala is supported by FPU21/05785 MIU.}
}

\author{
\IEEEauthorblockN{
Gonzalo Mancera\IEEEauthorrefmark{1},
Julian Fierrez\IEEEauthorrefmark{1},
Aythami Morales\IEEEauthorrefmark{1}\IEEEauthorrefmark{3},
Ruben Tolosana\IEEEauthorrefmark{1},
Daniel DeAlcala\IEEEauthorrefmark{1},
Francisco Jurado\IEEEauthorrefmark{2}
}

\IEEEauthorblockA{\IEEEauthorrefmark{1}BiometricsAI, Universidad Autonoma de Madrid, Spain}
\IEEEauthorblockA{\IEEEauthorrefmark{2}Computer Engineering Dept., Universidad Autonoma de Madrid, Spain}
\IEEEauthorblockA{\IEEEauthorrefmark{3}Mathematics Dept., Universidad de Las Palmas de Gran Canaria, Spain}
}

\maketitle

\begin{abstract}
This work investigates the application of the Membership Inference Test (MINT) to large language models (LLMs) fine-tuned with LoRA. MINT aims to determine whether specific data samples were used during model training, addressing key concerns related to data privacy, data leakage, and model transparency. This work builds on our previous LoRA-MINT framework based on perplexity, which addressed this same auditing problem under a strict black-box assumption, relying exclusively on the perplexity of candidate samples compared against a synthetic reference distribution in the domain. We propose an approach based on extracting intermediate representations from the model, capturing internal behaviors such as activations and parameter-dependent signals. These representations enable the construction of a MINT model that learns to distinguish between training and non-training samples, without requiring full access to the model parameters. We evaluated the method in 4 LLMs (Qwen, Phi-4, DeepSeek-V3, and LLaMA 3.2-3B) fine-tuned in three datasets covering medical, mathematical, and academic reasoning tasks. The results show that MINT is effective across different architectures and domains, achieving performance scores ranging from 95\% to 98\%, and demonstrating robustness under varying model configurations and data distributions. By allowing efficient auditing of whether sensitive records were used during fine-tuning without requiring full model access, the proposed MINT approach provides a practical mechanism for data provenance in accordance with emerging privacy and AI governance regulations in real-world LLM deployments. For the sake of concreteness and current relevance, our discussion and experiments are centered on LoRA-adjusted LLMs, but note that most of the presented methodology is easily applicable for auditing training data given any other technique for adapting LLMs or, more generally, any AI model.
\end{abstract}


\section{Introduction}
LLMs and VLMs are now revolutionizing multiple domains, including healthcare \cite{romero25food,ROMEROTAPIADOR2026111676}, law \cite{PENA-Layout,miguel25icdar}, human resources management \cite{mancera2025pba,pena2025addressing}, education \cite{2025_nature_education,irigoyen26k12}, and finance \cite{du2025natural}, significantly transforming the way textual information is processed and interpreted. These recent advances have enabled the analysis of large-scale textual corpora, accelerating progress in NLP applications. However, these models require vast amounts of data during training, and the provenance and public availability of such data remain controversial. 

Protecting sensitive information in AI applications \cite{2021_TPAMI_SensitiveNets_Morales,2022_AI_SensitiveLoss_IS} is now critical in order to preserve user trust and ensure compliance with increasingly stringent social concerns about the proper use of AI systems \cite{2023_SNCS_Human-Centric_Pena}. In response to these concerns, the European Union introduced new regulations in June 2024\footnote{\url{https://artificialintelligenceact.eu}}, aimed at governing the deployment of artificial intelligence systems \cite{irigoyen26risks}. This regulatory framework reflects a larger effort to address the societal and ethical implications of AI technologies, particularly in NLP applications.

A central objective of this legislation is the protection of fundamental rights through mechanisms that promote transparency and accountability in the design and use of AI systems. Consequently, both this framework and related international guidelines emphasize the need for auditing methodologies that enable the responsible deployment of AI/NLP technologies, while also supporting a better understanding of their behavior \cite{serna27unravel} and potential risks \cite{irigoyen26risks}.

In the context of increasing privacy concerns in AI deployments \cite{privacy2027}, a related work proposed Membership Inference Attacks (MIAs), which study whether an adversary can infer if specific data were used during model training \cite{shokri2017membership}. Previous work has shown that such attacks are feasible and has proposed a variety of mitigation and evaluation strategies.

Beyond traditional MIAs, recent work on privacy in machine learning has studied the possibility of inferring whether specific data samples were used during training, highlighting both the feasibility of such an inference and its implications for data protection and model leakage \cite{shokri2017membership,watson2021importance}. These approaches typically operate under adversarial assumptions, where an attacker exploits the behavior of the model to extract information about the training dataset.

\begin{figure}[t!]

\centering
\includegraphics[width=0.85\linewidth]{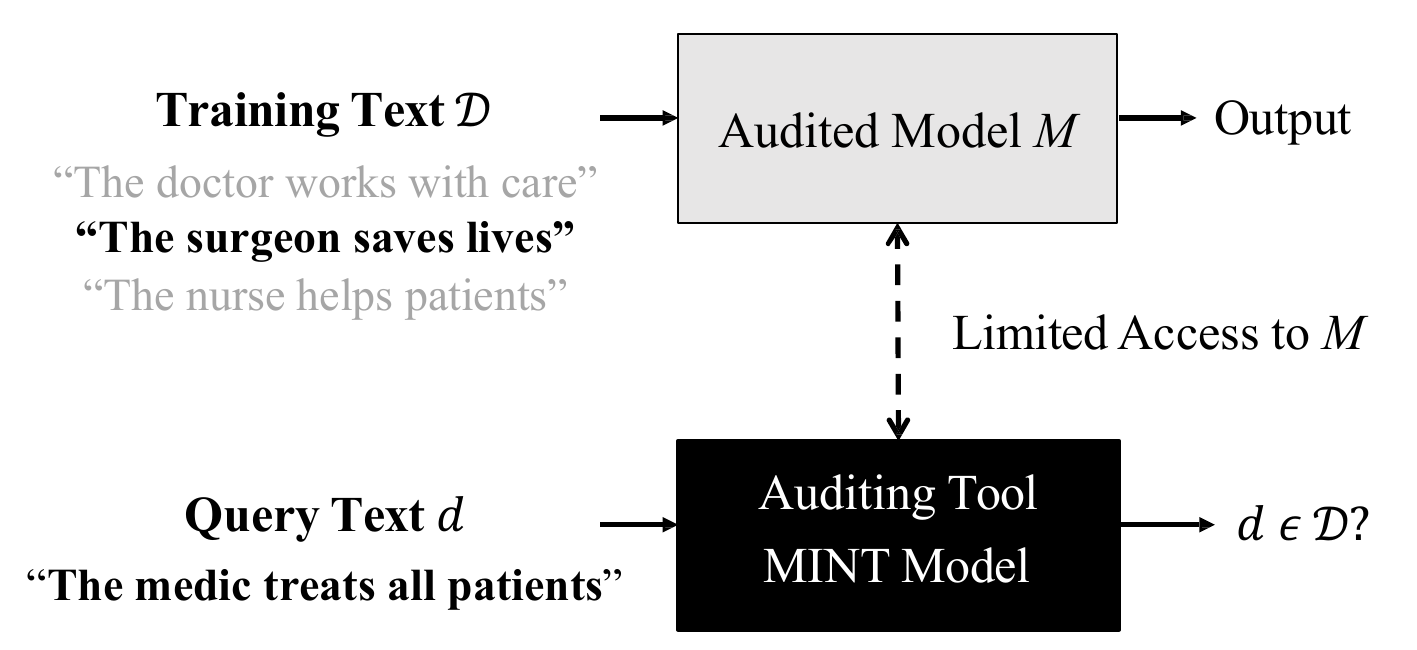}
\caption{MINT discerns whether given data (\textit{d}) were utilized in the training process of an AI Model (\textit{M}) trained with a specific database (\(\mathcal{D}\)).}
\label{fig:esquemamint}
\end{figure}

In line with recent research oriented to develop new technologies for auditing LLMs, the present paper addresses the challenge of identifying whether a given text was utilized in the training of a Language Model (see Fig.~\ref{fig:esquemamint}). The study aims to further highlight the versatility and applicability of the general MINT framework \cite{dealcala2024my} by adapting it to LoRA fine-tuned LLMs and demonstrating its effectiveness in language generation scenarios \cite{mancera2026auditing,dealcala2026my}. In this line, we extend our previous work LoRA-MINT~\cite{mancera2026auditing}, which relied solely on perplexity statistics computed from a synthetic reference distribution, by incorporating: 1) information from internal activations (`in' in inLoRA-MINT is from `internal'), 2) gradient-based sensitivity, and 3) output-level signals; trading a mild increase in required model access for consistently higher detection performance across the same model families and datasets.

The main contributions of this work are as follows.

\begin{itemize}
\item We adapt the MINT framework to LoRA fine-tuned LLMs using internal representations, enabling membership inference without full access to model parameters.

\item We evaluate 4 LoRA fine-tuned LLMs (\textbf{Qwen, Phi-4, DeepSeek-V3,} and \textbf{LLaMA 3.2-3B}) across three datasets (medical, mathematical, and academic), analyzing the effect of architecture, scale, and domain.

\item Results show that this representation-based approach effectively distinguishes training from non-training samples across models and domains.
\end{itemize}

For the sake of concreteness and current relevance, our discussion and experiments are centered on LoRA-adjusted LLMs, but note that the presented methodology is easily applicable for auditing training data given any other technique for adapting LLMs or, more generally, any AI model.

\section{Related Works}

\subsection{Membership Inference Test (MINT)}

MINT (Membership Inference Test) \cite{dealcala2024my} is a framework designed to evaluate whether a given machine learning model was trained or not on given data. Its main use as a data training auditing tool assumes some level of cooperation from the model provider, so that the development and implementation of the model follow legal and ethical requirements.

The primary goal of MINT is to establish reliable and broadly applicable evaluation procedures that can operate across multiple data modalities, including text, image, and audio domains. These procedures are designed to accommodate different degrees of transparency, depending on the amount of information disclosed by the model developer about the training process and the behavior of the system \cite{dealcala2024my,mancera2025my}. Consequently, the framework spans a wide spectrum of settings, ranging from highly opaque scenarios with limited or no disclosure to fully transparent environments where complete model details are available \cite{dealcala2024comprehensive,dealcala2025active}.

Within the MINT framework, auditors evaluate a model with full access to it or limited information about its architecture and training process. This makes MINT particularly relevant for practical compliance settings, especially in the context of evolving data protection regulations and legal requirements.

Unlike traditional Membership Inference Attacks (MIAs), which often rely on training shadow models to approximate the behavior of the target model, MINT operates directly on the original model. This leads to different MINT methodologies depending on the amount of information available from the  model being audited. As a result, the effectiveness of MINT varies depending on factors such as the level of access to the model and the type of data modality, making it adaptable to a wide range of audit scenarios \cite{dealcala2024comprehensive,gMINT2025,mancera2026membership,dealcala2025active}.

\subsection{Low-Rank Adaptation (LoRA)}
Low-Rank Adaptation (LoRA)~\cite{hu2022lora} is a parameter-efficient fine-tuning (PEFT) method that facilitates the specialization of massive pre-trained language models without necessitating a full gradient update of all network parameters. 

Instead of modifying the primary weight tensors, LoRA introduces a bottleneck structure consisting of two low-rank matrices that approximate the weight residuals within a significantly lower-dimensional manifold. Formally, for any static weight matrix $W_0 \in \mathbb{R}^{d \times k}$, LoRA keeps $W_0$ frozen and optimizes two auxiliary matrices $A \in \mathbb{R}^{d \times r}$ and $B \in \mathbb{R}^{r \times k}$, where the rank $r \ll \min(d, k)$. 

The final adapted weight is defined by:
\begin{equation}
W = W_0 + BA
\label{eq:W}
\end{equation}

This low-rank factorization drastically diminishes the volume of optimizeable parameters—frequently by multiple orders of magnitude—without compromising the model's effectiveness on target tasks. By restricting gradient updates solely to LoRA adapters, the foundational weights remain unchanged, which facilitates efficient storage, modularity, and rapid switching between different tasks.

Furthermore, LoRA fundamentally alters the optimization landscape during the fine-tuning process, which may impact the mechanisms of data memorization and internal representation. This unique property makes LoRA-tuned architectures a compelling subject for \textit{MINT} and privacy-aware technologies \cite{2017_Access_HEmultiDTW_Marta,bioengineering13050511}, as the distinct partitioning between static and trainable components could influence how sensitive information is distributed throughout the model.
\begin{figure*}[t!]
\centering
\includegraphics[width=\textwidth]{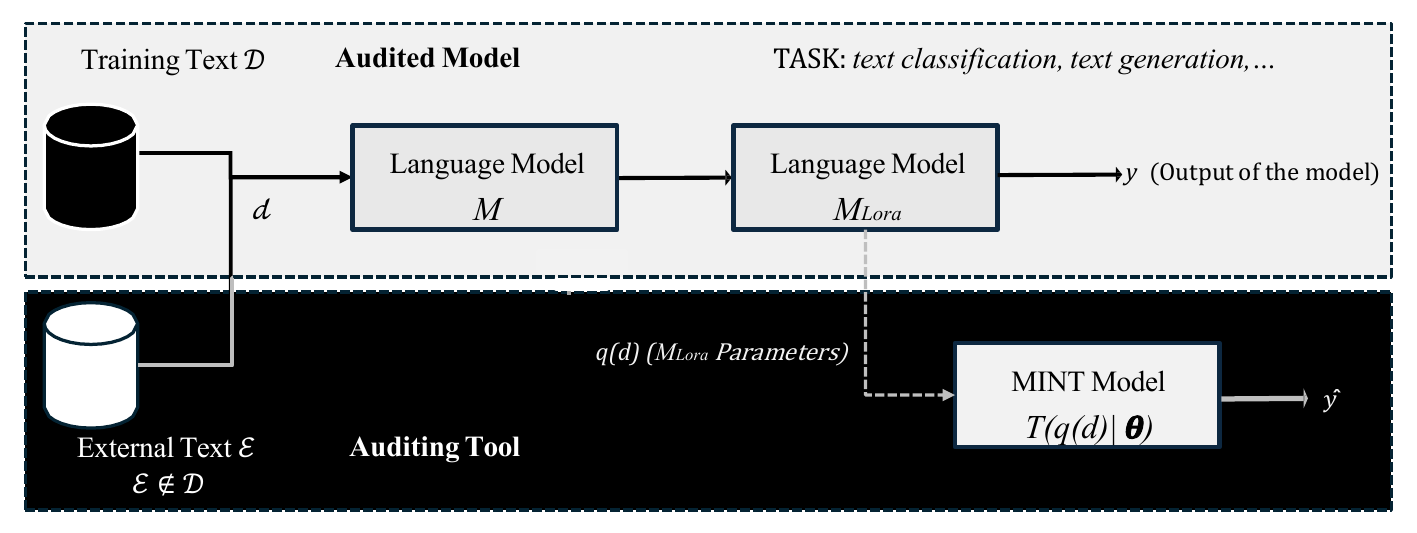}
\caption{The MINT model \( T(q(d) \mid \boldsymbol{\theta}) \) is trained to determine whether a given data instance $d$ was part of the training data $\mathcal{D}$ used to develop a language model $M_{\text{LoRA}}$. The inputs to the MINT model are represented by $q(d)$, which consists of features extracted from $M_{\text{LoRA}}$, including activation-based statistics, gradient-derived sensitivity measures, and output-level features computed with respect to the sample $d$.}
\label{fig:esquemamint_ext}
\end{figure*}
\section{Experimental Framework}

\subsection{Problem Formulation}

Given a training dataset $\mathcal{D}$ and an external dataset $\mathcal{E}$, we consider samples $d \in \mathcal{D} \cup \mathcal{E}$. We consider a base model $M$ trained for a specific task, which is further fine-tuned using LoRA, resulting in $M_{\text{LoRA}}$. For each input sample $d$, the model produces an output $y$ as a function of $d$ and the learned parameters $\boldsymbol{w}$, i.e.,
\[
y = M_{\text{LoRA}}(d \mid \boldsymbol{w}).
\]

In this setting, we assume access to intermediate representations extracted from the model, denoted as $q(d)$, which correspond to parameter-dependent signals such as activations, internal states, or subsets of model parameters computed when processing $d$. These representations capture how the model responds internally to different inputs.

We define the MINT model as a parametric function:
\[
T(q(d) \mid \boldsymbol{\theta}),
\]
where $\boldsymbol{\theta}$ are the learnable parameters of the model.

The MINT model is trained using representations $q(d)$ calculated on both datasets $\mathcal{D}$ and $\mathcal{E}$, ensuring that it captures differences in the internal behavior of the model for training and non-training samples.

The MINT model then predicts membership as follows:

\[
\hat{y} = T(q(d) \mid \boldsymbol{\theta}) = \begin{cases}
1 & \text{if } d \in \mathcal{D}, \\
0 & \text{if } d \in \mathcal{E}.
\end{cases}
\]

This formulation relies on the use of partial parameter-level information $q(d)$ associated with each
sample to distinguish between training and non-training data. This setting assumes a
certain degree of collaboration from the model developer, as access to this parameter-level information
is required.

As a result, this approach aligns with the principles of transparency and trustworthiness, enabling controlled
auditing mechanisms in scenarios where developers expose only minimal and carefully selected internal
signals. This limited disclosure is designed to avoid compromising the integrity, performance, or
confidentiality of the underlying model while still supporting an effective audit.

The method can be summarized as a pipeline:
(i) construct a reference set by collecting parameter-level representations $q(d)$ from samples with known membership labels,
(ii) use these representations to train the MINT model $T(\cdot \mid \boldsymbol{\theta})$ as a binary classifier,
(iii) extract the corresponding representations $q(d)$ for each candidate sample $d \in \mathcal{C}$, and
(iv) predict membership by evaluating $T(q(d) \mid \boldsymbol{\theta})$.

\section{Experimental Framework}
We begin by introducing the text datasets used in this study, which consist of three datasets covering instruction-based text generation and reasoning tasks across different domains. We then describe the 4 LLMs selected for auditing, detailing their main characteristics and configurations. Finally, we present the experimental protocol followed in this work.
\subsection{Databases}\label{databases}

\begin{itemize}

\item \textit{Medical-o1-reasoning-SFT} \cite{chen2024huatuogpto1medicalcomplexreasoning}. A medical reasoning dataset with verifiable question–answer pairs enriched with structured reasoning trajectories, supporting diagnostic and analytical reasoning in healthcare scenarios.

\item \textit{CAMEL Math Reasoning} \cite{li2023camel}. A synthetic multi-agent dataset built from role-playing agent dialogs, emphasizing step-by-step problem-solving across mathematical tasks.

\item \textit{Maths \& College Reasoning} \cite{ajibawa2023mathscollege}. A college-level dataset spanning algebra, calculus, probability, statistics, and differential equations, formatted as instruction–response pairs to evaluate multi-step reasoning.

\end{itemize}

\subsection{Audited Large Language Models} \label{audited}

Different models $M$ are evaluated to analyze whether variations in the architecture, scale, and complexity of the audited model affect the performance of MINT. In this study, we consider a set of LLMs that have been fine-tuned using LoRA, denoted $M_{\text{LoRA}}$, which allows efficient adaptation while preserving the capabilities of the base model. The fine-tuning process is performed using the datasets described in Section \ref{databases}, ensuring consistency between experimental conditions.

The selected models are chosen to provide a diverse and representative comparison between recent large language model families. These include \textbf{Qwen} \cite{bai2023qwen}, \textbf{Phi-4} \cite{abdin2024phi}, \textbf{DeepSeek-V3} \cite{liu2024deepseek}, and \textbf{LLaMA 3.2-3b} \cite{grattafiori2024llama}. Together, these models span different design choices in terms of scale, training data, and architectural refinements.

\begin{table*}[h!]
\centering
\caption{Performance comparison (AUC) of different MINT-based variants across three datasets and 4 LoRA fine-tuned models. Layer-based configurations are reported as Antepenultimate (L1), Penultimate (L2), and Last Layer (L3).}
\begin{tabular}{lllcccc}
\hline
\textit{DATASET} & \textsf{PROPOSED METHOD} &  & \textbf{Qwen3} & \textbf{Phi-4} & \textbf{DeepSeek-V3} & \textbf{LLaMA 3.2-3b} \\ \hline
\multicolumn{1}{l|}{\multirow{9}{*}{\textit{Medical-o1-reasoning-SFT}}}   
& Loss\cite{liu2022membership} (Relevant literature 2022) &  & 0.7508 & 0.9390 & 0.8840 & 0.8000 \\
\multicolumn{1}{l|}{} 
& MinK \cite{shi2023detecting} (Relevant literature 2024) &  & 0.7690 & 0.9440 & 0.8840 & 0.7770 \\
\multicolumn{1}{l|}{} 
& MinK++\cite{zhang2024min} (Relevant literature 2025) &  & 0.7780 & 0.9590 & 0.8625 & 0.7812 \\
\multicolumn{1}{l|}{} 
& \textsf{Antepenultimate Layer (L1)} &  & 0.8306 & 0.9800 & 0.9697 & 0.8811 \\
\multicolumn{1}{l|}{} 
& \textsf{Penultimate Layer (L2)} &  & 0.9179 & 0.9788 & 0.9840 & 0.9202 \\
\multicolumn{1}{l|}{} 
& \textsf{Last Layer (L3)} &  & 0.9272 & 0.9811 & 0.9846 & 0.9210 \\
\multicolumn{1}{l|}{} 
& \textsf{OnlyGrad} &  & 0.6706 & 0.7132 & 0.6683 & 0.5927 \\
\multicolumn{1}{l|}{} 
& \textsf{GradProbLoss} &  & 0.7920 & 0.6323 & 0.9435 & 0.8394 \\
\multicolumn{1}{l|}{} 
& \textsf{\underline{All3Layers}} &  & \underline{0.9302} & \underline{0.9819} & \underline{0.9851} & \underline{0.9240} \\ \hline

\multicolumn{1}{l|}{\multirow{9}{*}{\textit{CAMEL Math Reasoning}}}      
& Loss\cite{liu2022membership} (Relevant literature 2022) &  & 0.8838 & 0.9160 & 0.9051 & 0.8090 \\
\multicolumn{1}{l|}{} 
& MinK\cite{shi2023detecting} (Relevant literature 2024) &  & 0.8720 & 0.9160 & 0.9050 & 0.8090 \\
\multicolumn{1}{l|}{} 
& MinK++\cite{zhang2024min} (Relevant literature 2025) &  & 0.7823 & 0.8730 & 0.8465 & 0.7985 \\
\multicolumn{1}{l|}{} 
& \textsf{Antepenultimate Layer (L1)} &  & 0.9294 & 0.9533 & 0.9841 & 0.8948 \\
\multicolumn{1}{l|}{} 
& \textsf{Penultimate Layer (L2)} &  & 0.9429 & 0.9496 & 0.9865 & 0.9591 \\
\multicolumn{1}{l|}{} 
& \textsf{Last Layer (L3)} &  & 0.9485 & 0.9480 & 0.9843 & 0.9630 \\
\multicolumn{1}{l|}{} 
& \textsf{OnlyGrad} &  & 0.7073 & 0.8047 & 0.7895 & 0.7536 \\
\multicolumn{1}{l|}{} 
& \textsf{GradProbLoss} &  & 0.9201 & 0.9400 & 0.9254 & 0.8639 \\
\multicolumn{1}{l|}{} 
& \textsf{\underline{All3Layers}} &  & \underline{0.9514} & \underline{0.9578} & \underline{0.9901} & \underline{0.9645} \\ \hline

\multicolumn{1}{l|}{\multirow{9}{*}{\textit{Maths \& College Reasoning}}} 
& Loss\cite{liu2022membership} (Relevant literature 2022) &  & 0.9380 & 0.9210 & 0.9610 & 0.7850 \\
\multicolumn{1}{l|}{} 
& MinK\cite{shi2023detecting} (Relevant literature 2025) &  & 0.9440 & 0.9021 & 0.9840 & 0.8490 \\
\multicolumn{1}{l|}{} 
& MinK++\cite{zhang2024min} (Relevant literature 20245 &  & 0.9590 & 0.8464 & 0.8945 & 0.7954 \\
\multicolumn{1}{l|}{} 
& \textsf{Antepenultimate Layer (L1)} &  & 0.9848 & 0.9521 & 0.9810 & 0.9747 \\
\multicolumn{1}{l|}{} 
& \textsf{Penultimate Layer (L2)} &  & 0.9855 & 0.9801 & 0.9845 & 0.9743 \\
\multicolumn{1}{l|}{} 
& \textsf{Last Layer (L3)} &  & 0.9810 & 0.9821 & 0.9866 & 0.9726 \\
\multicolumn{1}{l|}{} 
& \textsf{OnlyGrad} &  & 0.6548 & 0.8424 & 0.7462 & 0.8420 \\
\multicolumn{1}{l|}{} 
& \textsf{GradProbLoss} &  & 0.9779 & 0.9398 & 0.9785 & 0.9627 \\
\multicolumn{1}{l|}{} 
& \textsf{\underline{All3Layers}} &  & \underline{0.9857} & \underline{0.9845} & \underline{0.9901} & \underline{0.9792} \\ \hline
\end{tabular}

\label{tab:my-table}
\end{table*}
\subsection{Experimental Protocol}
The audited model $M_{\text{LoRA}}$ is obtained by fine-tuning a base language model $M$ using Low-Rank Adaptation (LoRA) on a training set of approximately $13$k samples. The LoRA configuration employs a rank of $r=32$, a scaling factor $\alpha=32$, and a dropout rate of $0.15$, introducing trainable low-rank updates into selected transformer modules (including query, key, value, output, and feed-forward projections) while preserving the original model parameters. The fine-tuning process is carried out for 10 epochs using an AdamW-based optimizer with a learning rate of $10^{-4}$.

To characterize the behavior of the audited model $M_{\text{LoRA}}$ with respect to each input sample $d$, we extract a fixed-dimensional feature representation that captures multiple aspects of the model's internal and output responses.

First, we calculate the statistical descriptors from the internal activations of the last three layers of $M_{\text{LoRA}}$. Let $h^{(l)}(d) \in \mathbb{R}^{T \times D}$ denote the hidden representation in layer $l$ when processing sample $d$, where $T$ is the sequence length and $D$ is the hidden dimension. For each selected layer, we derive 4 summary statistics based on the token-averaged representation: 1) the mean value in all feature dimensions, 2) the standard deviation in all feature dimensions, 3) the $\ell_2$ norm of the averaged representation, and 4) the average $\ell_2$ norm computed in all token-level representations. These features capture both the global magnitude and the variability of internal activations. This results in $4$ features per layer and a total of $12$ activation-based features.

Second, we incorporate a sensitivity-related characteristic computed as the $\ell_2$ norm of the gradient with respect to the input embeddings of $M_{\text{LoRA}}$, resulting in a single scalar.

Finally, we include output-based features obtained from $M_{\text{LoRA}}(d \mid \boldsymbol{w})$, namely prediction loss and a confidence score calculated as the maximum predicted probability averaged between tokens. These contribute two additional features.

In general, each input sample $d$ is represented by a dimension vector of characteristics $15$, combining internal activation statistics ($12$), a gradient-based sensitivity measure ($1$) and the output signal ($2$). This $15$-dimensional representation is obtained in the experimental setting where features are extracted from the last three layers of $M_{\text{LoRA}}$, producing $4$ statistical descriptors per layer ($4 \times 3 = 12$). This enables a comprehensive characterization of the model for subsequent membership inference.

The MINT model is a fully connected neural network that operates on the extracted feature vectors. The input dimensionality depends on the feature configuration used in each experimental setting: it is $15$ when using the full representation based on the last three layers of $M_{\text{LoRA}}$, $7$ when using a single-layer configuration and $1$ when using individual signals such as gradient, loss, or probability features independently. The network consists of two hidden layers with $32$ and $16$ neurons, respectively. Each layer is followed by batch normalization, a SiLU activation function, and dropout regularization with rates of $0.2$ and $0.1$, respectively. The final layer outputs a single scalar value corresponding to the predicted membership score.

The model is trained using a binary cross-entropy loss with logits for stable binary classification. The optimization is performed using the AdamW optimizer with a learning rate of $3 \times 10^{-4}$ and a weight decay of $10^{-4}$, and the training process is carried out over $30$ epochs.

\section{Experiments and Results}

To systematically evaluate the effectiveness of different feature configurations for membership inference, we design six experimental settings that progressively incorporate deeper representations extracted from the audited model \( M_{\text{LoRA}} \). Experiments are conducted on three datasets (\textit{Medical-o1-reasoning-SFT}, \textit{CAMEL Math Reasoning}, and \textit{Maths \& College Reasoning}) and 4 LoRA fine-tuned models (\textbf{Qwen3, Phi-4, DeepSeek-V3}, and \textbf{LLaMA 3.2-3B}), using AUC as the evaluation metric. See Table~\ref{tab:my-table} for the results.

\begin{itemize}

\item \textsf{Experiment 1 (Antepenultimate Layer, L1)} uses only the statistical descriptors extracted from the third-to-last layer, resulting in a 7-dimensional feature vector, and achieves strong performance across all datasets (e.g., up to 0.9800 for \textbf{Phi-4} and 0.9841 for \textbf{DeepSeek-V3}), indicating that relatively shallow internal representations already contain relevant membership signals, although performance remains below deeper configurations. 

\item \textsf{Experiment 2 (Penultimate Layer, L2)} relies on features from the second-to-last layer, also with a 7-dimensional representation, and consistently improves over L1 (e.g., \textbf{Qwen3} improves from 0.8306 to 0.9179 on \textit{Medical-o1}), suggesting that intermediate representations provide a better balance between abstraction and memorization. 

\item \textsf{Experiment 3 (Last Layer, L3)} uses the final transformer layer, again with a 7-dimensional feature vector, and further improves performance in most cases (e.g., \textbf{DeepSeek-V3} reaches 0.9846 and 0.9843 on \textit{Medical} and \textit{CAMEL} datasets, respectively), confirming that deeper layers encode stronger membership-related signals.

\item \textsf{Experiment 4 (OnlyGrad)} evaluates a single gradient-based sensitivity feature (1-dimensional), resulting in significantly lower performance across all datasets (e.g. 0.6706 for \textbf{Qwen3} and 0.5927 for \textbf{LLaMA 3.2-3B} on \textit{Medical-o1}), indicating that gradients alone are insufficient to reliably capture membership information. 

\item \textsf{Experiment 5 (GradProbLoss)} combines gradient, loss, and confidence-based features into a compact 3-dimensional representation, yielding substantial improvements over the gradient-only setting (e.g. 0.9435 for \textbf{DeepSeek-V3} on \textit{Medical-o1} and 0.9779 for \textbf{Qwen3} on \textit{Maths \& College Reasoning}), demonstrating the complementarity of output-level and sensitivity signals.

\item \textsf{Experiment 6 (All3Layers)} integrates activation-based features of the antepenultimate, penultimate, and last layers together with gradient and output signals, resulting in a 15-dimensional feature vector and achieving the best overall performance across all datasets and models (e.g. 0.9851 for \textbf{DeepSeek-V3} on \textit{Medical-o1} and 0.9901 on both \textit{CAMEL} and \textit{Math} datasets), confirming that combining multi-layer internal representations with gradient and output features provides the most comprehensive characterization of model behavior for MINT.

\end{itemize}

\subsection{Ablation Study of LoRA Target Modules}

To quantify the impact of different injection points of LoRA within the transformer architecture, we conduct a structured ablation study with five configurations using the \textbf{Phi-4} model in the \textit{Medical-o1-SF} dataset. This setup enables us to analyze how attention- and MLP-based adaptations contribute to detection performance.

We begin by evaluating the adaptation to partial attention, where only \texttt{q\_proj} and \texttt{k\_proj} are enabled. We then extend this to full attention adaptation by activating \texttt{q\_proj}, \texttt{k\_proj}, \texttt{v\_proj}, and \texttt{o\_proj}. Next, we isolate the contribution of the MLP layers by enabling only \texttt{up\_proj} and \texttt{down\_proj}. This is followed by the full MLP configuration, where all three layers of MLP (\texttt{gate\_proj}, \texttt{up\_proj} and \texttt{down\_proj}) are activated. Finally, we evaluate the complete LoRA setup, which combines full attention with the full MLP (i.e., all three MLP layers), representing the maximum adaptation capacity.
\begin{table}[t!]
\centering

\caption{
Ablation study of LoRA target modules using the \textbf{Phi-4} model evaluated on the \textit{Medical-o1-SF} dataset. All experiments are conducted using the configuration that aggregates the representations from the last three transformer layers together with the MaxGradProb scoring criterion. Active LoRA injection points are shown in bold, while inactive modules appear in gray. Each configuration isolates a specific subset of attention or MLP pathways to quantify their contribution to detection performance.
}
\begin{tabular}{lll}
\hline
\multicolumn{1}{c}{Target Modules} & & \multicolumn{1}{c}{AUC} \\
\hline

\ttfamily 

% Row 1
\begin{tabular}[c]{@{}l@{}}
\textbf{q\_proj}, \textbf{k\_proj}, 
\textcolor{gray}{v\_proj}, \textcolor{gray}{o\_proj},\\
\textcolor{gray}{gate\_proj}, \textcolor{gray}{up\_proj}, \textcolor{gray}{down\_proj}
\end{tabular}
& & 0.6178 \\
\hline

\ttfamily 

% Row 2
\begin{tabular}[c]{@{}l@{}}
\textbf{q\_proj}, \textbf{k\_proj}, \textbf{v\_proj}, \textbf{o\_proj},\\
\textcolor{gray}{gate\_proj}, \textcolor{gray}{up\_proj}, \textcolor{gray}{down\_proj}
\end{tabular}
& & 0.6978 \\
\hline

\ttfamily 

% Row 3
\begin{tabular}[c]{@{}l@{}}
\textcolor{gray}{q\_proj}, \textcolor{gray}{k\_proj}, \textcolor{gray}{v\_proj}, \textcolor{gray}{o\_proj},\\
\textcolor{gray}{gate\_proj}, \textbf{up\_proj}, \textbf{down\_proj}
\end{tabular}
& & 0.7945 \\
\hline

\ttfamily 

% Row 4
\begin{tabular}[c]{@{}l@{}}
\textcolor{gray}{q\_proj}, \textcolor{gray}{k\_proj}, \textcolor{gray}{v\_proj}, \textcolor{gray}{o\_proj},\\
\textbf{gate\_proj}, \textbf{up\_proj}, \textbf{down\_proj}
\end{tabular}
& & 0.9019 \\
\hline

\ttfamily 

% Row 5
\begin{tabular}[c]{@{}l@{}}
\textbf{q\_proj}, \textbf{k\_proj}, \textbf{v\_proj}, \textbf{o\_proj},\\
\textbf{gate\_proj}, \textbf{up\_proj}, \textbf{down\_proj}
\end{tabular}
& & 0.9819 \\
\hline

\end{tabular}

\label{tab:lora_ablation}
\end{table}
The results, summarized in Table~\ref{tab:lora_ablation}, reveal a clear trend. Partial attention adaptation yields limited performance (AUC = 0.6178), while full attention provides a moderate improvement (AUC = 0.6978). In contrast, MLP-based adaptations contribute significantly more, with partial MLP reaching an AUC of 0.7945 and full MLP (i.e., all three layers enabled) achieving 0.9019. The best performance is obtained when combining both both attention and the full MLP stack, reaching an AUC of 0.9819.

In general, these findings indicate that MLP pathways—particularly when all three layers are enabled—play a more dominant role than attention modules in this setting, while their combination produces a strong synergistic effect.

\section{Conclusions}

We presented inLoRA-MINT, a feature-based framework for membership inference on LoRA fine-tuned LLMs. For the sake of concreteness and current relevance, our discussion and experiments are centered on LoRA-adjusted LLMs, but note that most of the presented methodology is easily applicable for auditing training data given any other technique for adapting LLMs or, more generally, any AI model. Our proposed approach inLoRA-MINT combines internal activation statistics, gradient-based sensitivity measures, and output-level signals to construct expressive representations of model behavior and then a binary membership classification.

Experiments on three benchmark datasets (\textit{Medical-o1-reasoning-SFT}, \textit{CAMEL Math Reasoning} and \textit{Maths \& College Reasoning}) and 4 state-of-the-art models (\textbf{Qwen3, Phi-4, DeepSeek-V3}, and \textbf{LLaMA 3.2-3B}) reveal consistent trends. Deeper transformer layers provide progressively stronger membership signals, with later layers outperforming earlier ones. Gradient-only features show limited effectiveness, while combining gradient, loss, and confidence signals leads to significant improvements, highlighting their complementarity.

The best performance is achieved by integrating activation statistics from the last three layers with gradient and output-based features, confirming that membership information is distributed across multiple levels of the model and that leveraging heterogeneous signals is essential for robust inference.

Finally, we observe differences between model families. Larger models such as \textbf{DeepSeek-V3} achieve higher and more stable AUC scores, suggesting stronger memorization effects, while smaller models such as \textbf{LLaMA 3.2-3B} show lower and more variable performance. \textbf{Qwen3} and \textbf{Phi-4} lie in an intermediate regime, indicating that model scale and architecture play a key role in privacy leakage.

In our future work, we will study and extend our training data auditing tools to evaluate other relevant risks \cite{irigoyen26risks} in multimodal LLMs (including VLMs \cite{dealcala2026my}). Analyzing biases \cite{2023_ECAIw_LFIT-XAI_Tello,pena2025addressing} and synthetic manipulations \cite{pavel25iccv} while maintaining privacy \cite{mancera2025pba} are also key topics for risk evaluation and mitigation in AI systems. Finally, we will apply these risk assessment and management principles to concrete domains such as document intelligence \cite{MUNOZHARO2026103969,miguel26icdar}, e-learning~\cite{irigoyen26k12,2025_nature_education}, gaming~\cite{daza2026ares}, and e-health~\cite{ROMEROTAPIADOR2026111676}.

\bibliographystyle{IEEEtran}
\bibliography{IEEEexample}

\end{document}